\documentclass[final]{cvpr}
\usepackage{url}
\usepackage{times}
\usepackage{epsfig}
\usepackage{graphicx}
\usepackage{amsmath}
\usepackage{amssymb}
\usepackage{bbm}
\usepackage[tight,footnotesize]{subfigure}
\usepackage[pagebackref=true,breaklinks=true,colorlinks,bookmarks=false]{hyperref}

\def\cvprPaperID{****} % *** Enter the CVPR Paper ID here
\def\confYear{CVPR 2021}

\begin{document}

%%%%%%%%% TITLE
\title{A Sub-band Approach to Deep Denoising Wavelet Networks and a Frequency-adaptive Loss for Perceptual Quality}

\author{Caglar Aytekin, Sakari Alenius, Dmytro Paliy and Juuso Gren\\
AAC Technologies\\
Itäinenkatu 11, 33210 Tampere\\
{\tt\small caglaraytekin@aactechnologies.com}
%\and
}

\maketitle

%%%%%%%%% ABSTRACT
\begin{abstract}
   In this paper, we propose two contributions to neural network based denoising.
   First, we propose applying separate convolutional layers to each sub-band of discrete wavelet transform (DWT)
   as opposed to the common usage of DWT which concatenates all sub-bands and applies a single convolution layer.
   We show that our approach to using DWT in neural networks improves the accuracy notably, due to keeping the sub-band order uncorrupted prior to inverse DWT.
   Our second contribution is a denoising loss based on top k-percent of errors in frequency domain.
   A neural network trained with this loss, adaptively focuses on frequencies that it fails to recover the most in each iteration.
   We show that this loss results into better perceptual quality by providing an image that is more balanced in terms of the errors in frequency components.
\end{abstract}

%%%%%%%%% BODY TEXT
\section{Introduction}

Image denoising is a fundamental part of any image signal processing (ISP) pipeline, and remains to be a challenging problem.
As the noise is present in any sensor, denoising is an inevitable processing step applied on RAW images in order to obtain a clean image.
Being usually the first pre-processing step, other processing steps depend on the performance of the denoising module.
Therefore, one might argue that the denoising part is the single most important part of an ISP.

Over the years, many approaches have been applied to the denoising problem. 
Traditional approaches to image denoising , such as NLM \cite{NLM}, K-SVD \cite{KSVD}, BM3D \cite{BM3D}, WNNM \cite{WNNM}, EPLL \cite{EPLL}, may suffer from poor performance, complex optimizations during test-time and the need to manually tune parameters.  
This is evident from run-time speeds and evaluation metrics of these methods in a benchmark \cite{SIDDBENCH}.

Recently, with the rise of deep learning, convolutional neural network (CNN) based denoising algorithms \cite{DIDU,DCHN,GRDN,BAUG,MSRD,DSGW,SKNAS} have been demonstrated to outperform the traditional methods. 
The main advantage of deep learning based methods over traditional methods is the end-to-end learning paradigm which enables avoiding some common problems in traditional methods such as hand-crafted ad-hoc regularization terms and run-time parameter tuning.
Moreover, although training of deep networks is rather complex, test-time does not suffer from complex optimization time per image, and is rather fast due to one single forward processing.

The end-to-end learning paradigm is most effective when the input and expected output of the method is clearly defined.
Earlier denoising datasets \cite{RENOIR,TID} obtained so-called noise-free images by capturing scenes with high-quality cameras with low ISO and long exposure. 
Noisy images are then obtained by adding simulated signal-dependent Gaussian noise.
However, for images obtained by smart-phones, even the best camera settings are not sufficient to obtain a clean, noise-free images as in high-quality DSLRs.
To address this difficulty, recently SIDD \cite{SIDD} was introduced, where a systematic procedure was applied for obtaining noise-free images and their noisy counterparts from smart-phone images.
SIDD provides excellent input,output pairs in order to train neural-networks to fully exploit the power of end-to-end learning with real data.

In order to assess the performance of recent denoising methods, NTIRE2019 \cite{NTIRE2019} and NTIRE2020 \cite{NTIRE2020} denoising challenges have been organized using SIDD as the challenge dataset.
Among the deep learning based denoising methods tested in these challenges, top-performing methods employ some common ideas.
From the architectural perspective, most popular approaches to denoising CNN utilize U-Net \cite{UNET}, ResNet \cite{RESNET} and DenseNet \cite{DENSENET} combinations. 
For example, residual dense block is a popular bottleneck structure and has been used in \cite{DCHN,GRDN,DSGW}.
Another increasingly popular idea is to utilize Discrete Wavelet Transform (DWT) within the neural networks as in \cite{DSGW} and by Huawei-team in \cite{NTIRE2020}.
Finally, both for convergence and performance, L1 loss is adopted in a large majority of the top-performing methods.

In this paper we propose two contributions to these common adoptions as follows.
\begin{itemize}
  \item First, we argue the presence of a sub-optimality in the common usage of DWT in neural networks. We propose to apply separate convolutional layers to each DWT sub-band and show that this approach outperforms the common usage.
  \item Second, we propose to apply L1 loss on discrete cosine transform frequency components which have highest errors (top k$\%$).
We show that this loss helps achieving visually more pleasing images, while keeping the overall distortion metric at a reasonable level.
\end{itemize}

%-------------------------------------------------------------------------
\section{Related Work}

In this section, we will shortly review some of the concepts that are related to the contributions of our proposed method.

\subsection{Discrete Wavelet Transform}
\label{DWT_theory}
In an $nxn$ discrete wavelet transform (DWT), a total of $n^2$ orthogonal wavelets are used to convolve an image and convolution results are downsampled by $n$.
For example, in a $2x2$ DWT, $4$ wavelets are usually denoted as $f_{LL}, f_{LH}, f_{HL}, f_{HH}$, where $L$ and $H$ denote low and high frequencies respectively.
Then from a signal $x$, a sub-band is obtained as follows. 
\begin{equation}
x_i=(x \circledast  f_{i}) \downarrow _2 , i \in \{LL,LH,HL,HH\}
\label{dwt_eq}
\end{equation}
Due to the orthogonality property of the wavelets, the original signal $x$ can be recovered by an inverse discrete wavelet transform (IDWT).

Discrete wavelet transform (DWT) has been used for denoising in both traditional \cite{TDWT} and deep methods \cite{MLWCNN,DSGW}.
A general advantage of DWT over Fourier Transform is its ability to keep both spatial and frequency information in the sub-bands.

For deep learning methods, DWT has another advantage: DWT both helps to increase the receptive field (due to downsampling) and at the same time prevents information loss (due to its invertible nature).
In \cite{MLWCNN}, DWT was utilized in a multi-level manner, where sub-bands after a DWT was concatenated in channel-dimension and passed through a convolutional layer. Resulting tensor is passed through another DWT and above process was repeated until desired level is achieved.
In the inverse procedure, output of a convolutional layer is de-concatenated channel-wise and given to IDWT and this is repeated until the dimensions are back to image-level.
In \cite{DSGW} a densely self-guided wavelet network was proposed that utilizes DWT following the same main approach as in \cite{MLWCNN}.

All of the approaches utilizing DWT in neural networks, concatenate sub-bands in channel dimension, run convolutional layers, de-concatenate and then apply inverse DWT (IDWT) to go back to original image domain.
We argue that the sub-band order information necessary for IDWT, may be lost after applying a convolution on concatenated sub-bands.
Therefore, in Section \ref{sect_DWT}, we propose to apply separate convolutional layers to each sub-band, thus keeping sub-band order information for IDWT.
We find that this approach outperforms the common one.

\subsection{Denoising Loss}
L1 Loss is the most commonly used loss among deep learning based denoising methods \cite{NTIRE2019,NTIRE2020}, as it has been shown to outperform L2 loss in terms of performance and convergence.
Some works in \cite{NTIRE2019,NTIRE2020} have also employed a joint loss by combining L1 and MS-SSIM in order to achieve good results in both evaluations.
Although distortion metrics such as L1, L2 and MS-SSIM are useful to objectively evaluate methods, they do not necessarily correlate well with visual quality which is the most important in the real use case. 
Especially, camera manufacturers have to take the perceptual quality into account as that is the single most important evaluation in terms of customer satisfaction.

Not only distortion and perception quality do not correlate well, moreover it was found in \cite{PERDIS} that they actually go against each other in a non-invertible problem setting such as denoising.
In other words, for a non-invertible problem there is a performance boundary that one cannot go beyond due to the nature of the problem, and among this boundary one can go either in favor of perception or distortion, and the other will suffer.

Although they are shown to outperform L2 also in terms of visual quality, L1 and MS-SSIM are still distortion metrics.
In order to obtain a better visual quality, in Section \ref{DCTLoss}, we propose an adaptive frequency loss that is dynamically changing according to frequency characteristics of image content and the performance of the neural network.

\begin{figure*}
  \centering \includegraphics[scale=0.8]{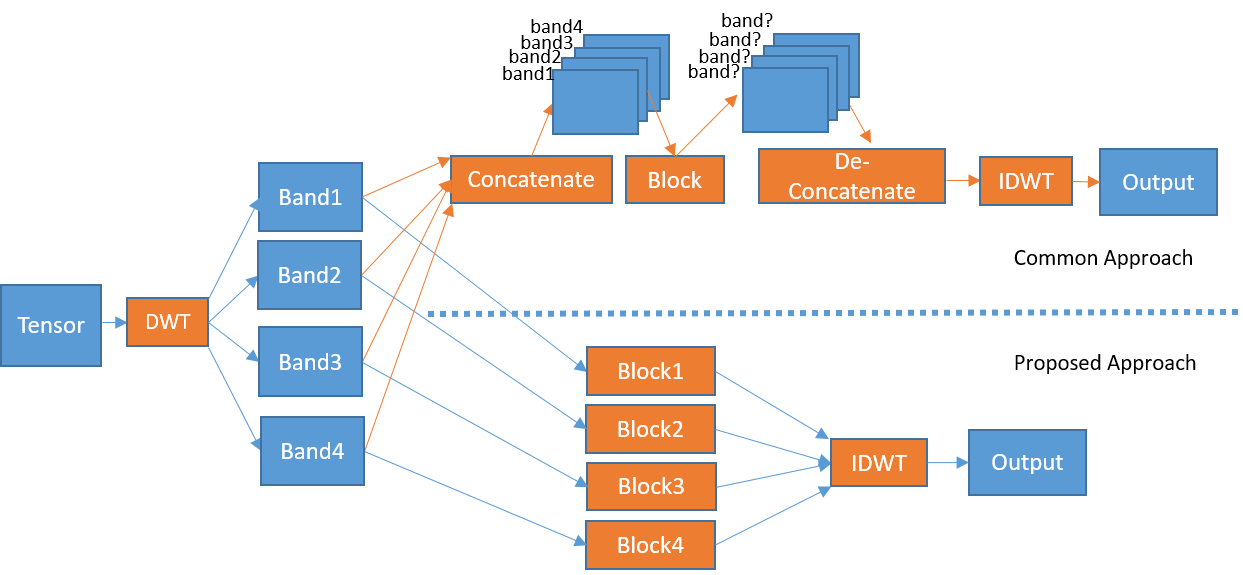}
  \caption{Common usage of Discrete Wavelet Transform (DWT) (up) and our approach (down)}
  \label{fig1}
\end{figure*}

\subsection{Residual Dense Block}

A residual dense block (RDB) combine properties of DenseNets and ResNets and was first proposed in a deep image super-resolution work.
In an RDB bottleneck, several convolutional layers are densely connected and usually there is a single convolution at the end in order to convert to input channel dimension so that input and output can be added as in ResNets. 
In \cite{RDB}, RDB was proposed for image super-resolution and denoising tasks as a contiguous memory mechanism and was shown to outperform compared bottlenecks.
In \cite{DCHN}, RDB was used as the main bottleneck for a modified U-Net based denoising task.
In \cite{DSGW}, RDBs were used in a self-guided wavelet network for image denoising.
In this paper we also choose RDB as our bottleneck, but we adapt the structure of RDB according to our proposed sub-band specialized convolutions as explained in Section \ref{sect_BOT}.

\section{Proposed Method}
\label{PropMet}

\subsection{How to Use Discrete Wavelet Transform in Neural Networks} \label{sect_DWT}
As discussed in Section \ref{DWT_theory}, image restoration neural networks that are using DWT, first apply the transform and then concatenate the generated sub-bands to a single tensor.
Then, operations are performed on this single tensor.
These operations can be convolutional blocks and for the sake of convenience, are simply denoted by "blocks" in Fig. \ref{fig1}.
Although the common usage of DWT in neural networks seems natural, we notice a potential problem which is explained as follows.
The concatenation operation and subsequent convolutional blocks could potentially disrupt the ordering of the sub-bands as illustrated in upper branch of Fig. \ref{fig1}.
Yet, for the de-concatenation which is necessary for IDWT -- and thus going back to the original domain prior to DWT --, the sub-band ordering has to be preserved.
Disrupting this ordering highly contradicts to the original motivation of using DWT: its invertible nature.
We see this as a potential problem in learning, as a neural network is expected to learn also the ordering information in the convolutional filters during training.

In order to remedy this situation, we propose to simply apply separate convolutional blocks per each sub-band as shown in lower branch of Fig. \ref{fig1}.
Separately applying convolutional layers to sub-bands discards inter-band communication, for example lower frequencies are processed separately without considering higher frequencies at all.
Although this might seem like another potential problem, we avoid it with a simple trick in a specially designed bottleneck in Section \ref{sect_BOT}.

\begin{figure}
  \centering \includegraphics[scale=0.9]{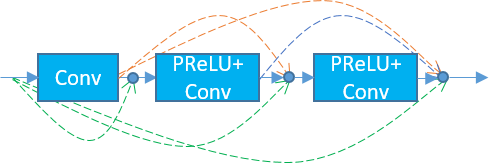}
  \caption{Dense Block (DB)}
  \label{fig2}
\end{figure}

\subsection{Residual Sub-band Dense Block} \label{sect_BOT}
In this section, we define the bottleneck block used in our method.
First, we review Residual Dense Blocks (RDBs) as we will use a similar bottleneck.
RDBs are widely used in many image restoration problems \cite{RDB,DCHN,DSGW}.
In RDB, the output of each operation (feature) is concatenated to the subsequent features as in DenseNets.
At the end, there is a single convolution that operates on concatenation of all features.
Moreover, the input to RDB is added to the output as in ResNets. 
In this work, we are going to use a variant of RDB utilizing our DWT approach, therefore we first take the dense block (DB) part which is RDB without the skip connection and is illustrated in Fig. \ref{fig2}.
\begin{figure}
  \centering \includegraphics[scale=0.8]{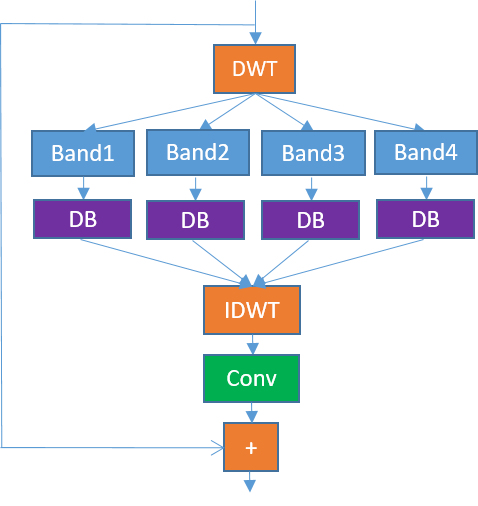}
  \caption{Residual Sub-band Dense Block (RSDB)}
  \label{fig3}
\end{figure}

\begin{figure*}
  \centering \includegraphics[scale=0.8]{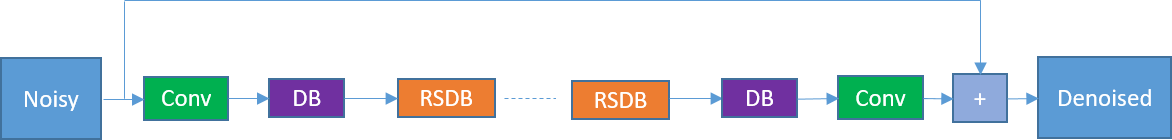}
  \caption{Neural Network Architecture}
  \label{fig4}
\end{figure*}

We will transfer the last convolution and skip connection in RDB to our bottleneck block which is defined next.
For convenience we will address our bottleneck as Residual Sub-band Dense Block (RSDB) block.
DWT is applied on a tensor and 4 sub-bands are obtained. 
We assumed a DWT with 2x2 wavelets -- hence 4 sub-bands --, but the idea can be easily extended to larger wavelets.
Then, a dense block (DB) is applied on each sub-band. 
The output of DB for each sub-band is given to inverse DWT operation.
Finally, a convolution is applied and the result is added to the input of the block.
Besides converting densely concatenated features to input feature dimension, final convolution also addresses the no inter-band communication issue explained in Section \ref{sect_DWT}.
By applying IDWT and applying convolution afterwards, we make sure that the bottleneck also performs an operation on original domain, where all bands' information is present.
The Residual Sub-band Dense Block is illustrated in Fig. \ref{fig3}.

\subsection{Loss Function: Top k-percent Frequency Loss}
\label{DCTLoss}
As explained in \cite{PERDIS}, for non-invertible problems there is a trade-off between the perceptual quality and distortion metric (such as PSNR), when it comes to extreme ends.
L1 loss is the dominant loss function that is used in many deep image restoration works \cite{NTIRE2019,NTIRE2020}.
Although advantage of L1 over L2 is proven in many previous studies, L1 is still largely a PSNR oriented loss.

Here we propose a loss that largely keeps the PSNR accuracy and also increase the perceptual quality.
We first take the Discrete Cosine Transform (DCT) of both the ground truth $y_{true}$ and the prediction $y_{pred}$, in order to decompose them to their frequency-related coefficients. 
We chose DCT due to its computational simplicity. Also, DCT is commonly used in image compression as a rough measure of perceptual quality as in JPEG \cite{JPEG}.
Next, we take the absolute error between each frequency coefficient of $y_{true}$ and $y_{pred}$.
These errors are sorted in a descending order and the loss is calculated by taking the mean of the top $k\%$ in the sorted absolute errors as in Eq. \ref{eq1}.

\begin{equation}
\begin{aligned}
DCT_{error}=|DCT(y_{true})-DCT(y_{pred})| \\
loss_k=\frac{1}{|K|}\sum_{i\in K}DCT_{error}^{(i)}
\end{aligned}
\label{eq1}
\end{equation}

In Eq. \ref{eq1}, $K$ is the set of elements of $DCT_{error}$ that are in top $k\%$.
During training, the top $k\%$ frequency loss dynamically changes according to prediction at each iteration, since the frequencies that fall into top $k\%$ changes as the network learns.
This ensures that the network is  concentrated on the frequencies that it fails to restore the most.
For example, a common data imbalance scenario may be that the bias term (zero frequency) dominates the others in a dataset where flat-like areas are more common.
In such a case, a neural network with uniform contributions of DCT coefficients to loss function would dominantly be trained for recovering the zero-frequency term, and may easily discard not-so-populated frequencies. That would result in bad visual quality in some locations in an image.
In the same scenario, when using our loss, once a network learns the bias-term well enough, zero frequency would not pop up in the top $k\%$ frequency errors anymore, therefore the network would concentrate on recovering frequencies that it fails the most: Possibly the minority frequency components that are less populated in the data.

Using the top $k\%$ frequency loss may result in a slight decrease in the performance in terms of distortion, since the goal is not to reduce the overall error.
The goal is to reduce the error in a more balanced way in frequency domain in order to provide a better visual quality.

\subsection{Neural Network Architecture}
The entire neural network architecture is illustrated in Fig. \ref{fig4}. 
First, the input is transformed to a higher-dimensional space via consecutive convolutional and dense blocks.
Then, multiple RSDB blocks are applied and the features are transferred back to image dimensions via consecutive dense and convolutional blocks.
We also utilize a big skip connection between the input and the output of the neural network.

\section{Experimental Results}

\subsection{Compared Bottlenecks}
\begin{figure}
  \centering
  \subfigure[Common DWT usage]{\label{fig:5a}\includegraphics[width=0.2\textwidth]{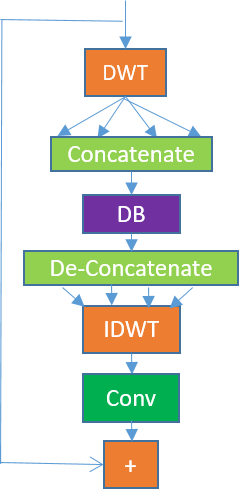}}\qquad
  \subfigure[No DWT]{\centering\label{fig:5b}\includegraphics[width=0.12\textwidth]{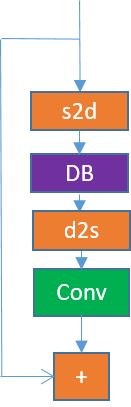}}
\caption{Compared Bottlenecks}
\label{fig5}
\end{figure}

In order to verify the effectiveness of our DWT approach to the common DWT approach and no-DWT approach, we have  conducted experiments with compatible blocks as shown in Fig. \ref{fig5}.
In particular, we have experimented on three blocks: (a) our proposed RSDB block shown in Fig. \ref{fig3}, (b) a compatible block based on common DWT approach as shown in Fig. \ref{fig:5a}, and (c) a compatible block where no DWT is applied as shown in Fig. \ref{fig:5b}. Note that for (c) we have used space-to-depth (s2d) and depth-to-space(d2s) operators in order to be compatible with downsampling and upsampling procedures in (a) and (b) due to DWT and IDWT.
The general neural network architecture for all experiments are exactly the same, only the bottleneck RSDB in Fig. \ref{fig4} is replaced with corresponding bottlenecks for each experiment.
The number of filters in compared bottlenecks are arranged slightly in order to keep the total number of parameters in the network similar.

\subsection{Experimental Setup and Details}
All of the neural networks used in this work were trained on original SIDD-Medium dataset with provided training and validation splits in \cite{NTIRE2019}.
Except for the last convolutional layer, we have always used the same number of filters in all convolutional layers (64). 
We have used 8 RSDB blocks in total.
We have trained all neural networks for 300 epochs in total, using RADAM \cite{RADAM} optimizer with initial learning rate of $0.0002$ and we have reduced the learning rate by 10 after 200 epochs. 
For the experiments that we have used top $k\%$ loss, we have gradually decreased $k$ by $1$ in each epoch, starting from $100$ to a minimum of $10$.
All networks are trained to process RGGB Bayer data, where 1-channel RGGB image is converted to 4-channel by space-to-depth operator as neural network input.
Neural network output is converted back to 1-channel RGGB image by depth-to-space operator.
All experiments use horizontal flip, vertical flip and rotation as data augmentation during training.
For all reported results in this paper, we have taken an ensemble of 8 results on all permutations of horizontal flip, vertical flip and rotation.
While doing the augmentation and the ensemble we have followed the approach of \cite{BAUG} in order to preserve the Bayer order for all samples, particularly cropping or padding respective locations to preserve RGGB Bayer order for the neural network.

\begin{figure}
  \centering \includegraphics[scale=0.5]{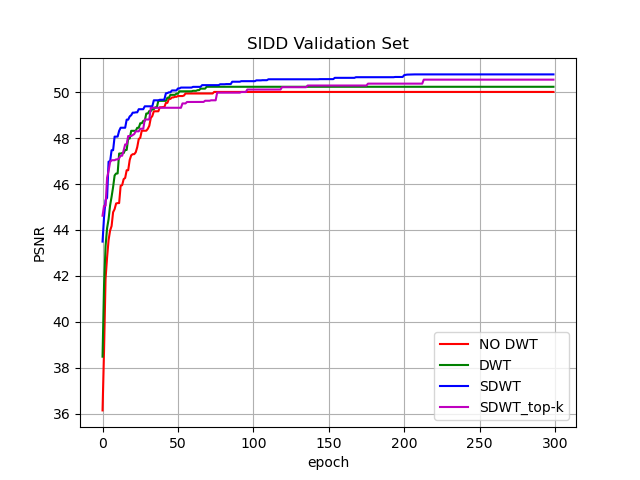}
  \caption{PSNR curves for compared methods on SIDD validation set: For a more clear illustration, we plot the curves so that a value at an epoch does not indicate validation PSNR at that epoch, but best validation PSNR obtained so far}
  \label{curves}
\end{figure}

\begin{figure*}
  \centering \includegraphics[scale=0.9]{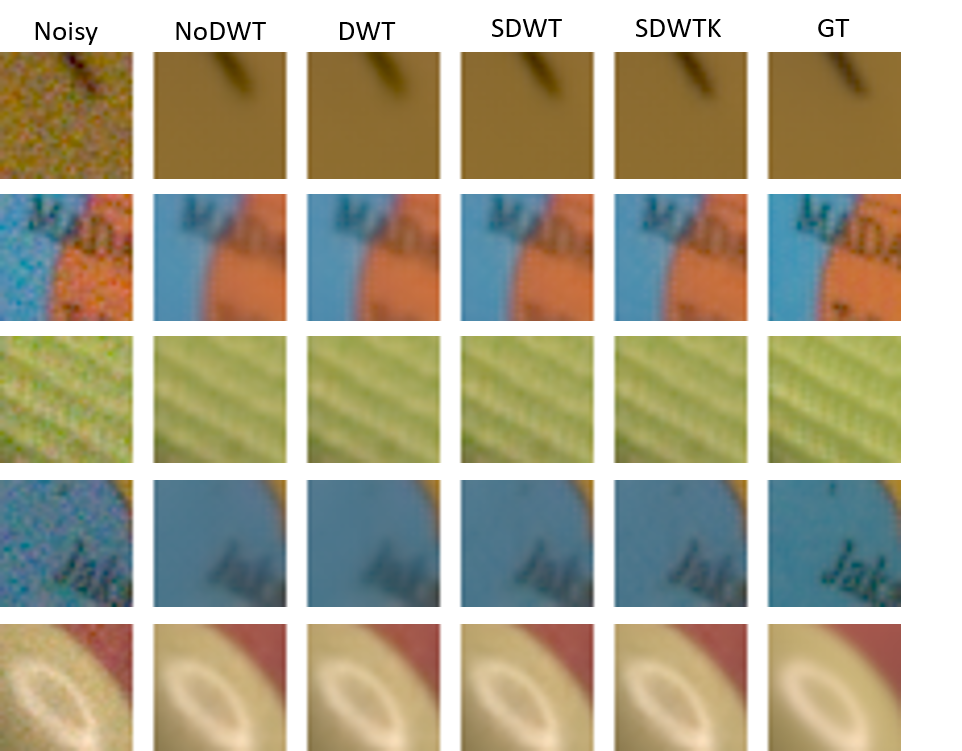}
  \caption{Visual Comparisons}
  \label{fig6}
\end{figure*}

\subsection{Comparison to Baseline}
\begin{table}
\caption{\label{siddp_val_psnr}PSNR values on SIDD+ validation datasets for methods.}

 \begin{tabular}{||c c c c c||} 
 \hline
 & No DWT & DWT & SDWT & $SDWT_{top-k}$ \\ [0.5ex] 
 \hline\hline
 SIDD+ & 52.38 & 52.40 & 52.56 & 52.46 \\
 \hline

\end{tabular}

\end{table}

In this section, we compare PSNR results on validation sets of SIDD and SIDD+ datasets of our method and the baselines.
For convenience, we call our method SDWT referring to separately applying operations on sub-bands (S) of DWT outputs.

We first analyze the training curves for compared algorithms in Fig. \ref{curves}.
It can be observed that our method learns faster at an already high PSNR in the beginning of the training compared to other methods.
Moreover as the training continues, SDWT generalization keeps getting better, whereas for others validation PSNR seems to saturate at lower values compared to SDWT.

At the time of implementing the method, we did not yet have access to SIDD+ dataset, therefore validation was made based on SIDD dataset.
Later, we had access to SIDD+ dataset too.
Hence, we share our results in SIDD+ validation set as well.
Similar to SIDD validation set, as it can be observed from Table \ref{siddp_val_psnr}, our SDWT outperforms the baselines in SIDD+ validation set too. 

As expected, the network trained with top-$k\%$ loss ($SDWT_{top-k}$ ) has slightly lower PSNR then SDWT.
Yet, as discussed earlier, the main advantage of this loss is perceptual quality.
Next, we share some images from SIDD+ validation dataset to visually compare methods.
After denoising, for better visual comparison, the images were processed by the ISP pipe provided by NTIRE-2020 Challenge organizers.

In Fig. \ref{fig6}, we show some enlarged crops from the SIDD+ validation dataset.
SDWT's advantage over other methods are clearly observed on all examples.
The perceptual improvement of top-$k\%$ loss ($SDWT_{top-k}$) over L1 loss (SDWT) is best observed in high-frequency areas where top-$k\%$ loss helps recovering the edges better.
The letters on images on rows 2 and 4 in Fig. \ref{fig6} are clear examples for this.

For further demonstration of the visual quality improvement that $SDWT_{top-k}$ brings, we have conducted experiments on our own modules in Section \ref{own_exps}, where the improvement is more evident.

\begin{table*}
\caption{\label{test_psnr}Comparison to state-of-the-art on SIDD+ test set.}

 \begin{tabular}{||c c c c c c||} 
 \hline
 Method & PSNR & SSIM & Time (s/MP) & GPU & Ensemble \\ [0.5ex] 
 \hline\hline
 Baidu Research Vision & 57.44 & 0.99789 & 5.76 & Tesla V100 & flip/transpose (x8) \\
 \hline
 HITVPC\&HUAWEI & 57.43 & 0.99788 & ? & GTX 1080 Ti & flip/rotate (x8) \\
 \hline
 Eraser 1 & 57.33 & 0.99788 & 36.50 & TITAN V & flip/rotate (x8) \\
 \hline
 $SDWT_{12,96,1}$ & 57.31 & 0.99784 & 6.63 & QUADRO RTX 6000 & flip/rotate (x8) \\
 \hline
 Samsung\_SLSI\_MSL & 57.29 & 0.99790 & 50 & Tesla V100 & flip/transpose (x8), models (x3) \\
 \hline
 Tyan & 57.23 & 0.99788 & 0.38 & GTX 1080 Ti & flip/transpose (x8), models (x3) \\
 \hline
 NJU-IITJ & 57.22 & 0.99784 & 3.5 & Tesla V100 & models (x8) \\
 \hline
 Panda & 57.20 & 0.99784 & 2.72 & GTX 2080 Ti & flip/rotate (x8), models (x3) \\
 \hline
 BOE-IOT-AIBD & 57.19 & 0.99784 & 0.61 & Tesla P100 & None \\
 \hline
 TCL Research Europe & 57.11 & 0.99788 & ? &RTX 2080 Ti & flip/rotate (x8), models (x3-5) \\
 \hline
 $SDWT_{8,64,1}$ & 57.10 & 0.99773 & 3.30 & QUADRO RTX 6000 & flip/rotate (x8) \\ 
 \hline
 Eraser 3 & 57.03 & 0.99779 & 0.31 & ? & ? \\
 \hline
 EWHA-AIBI & 57.01 & 0.99781 & 55 & Tesla V100 & flip/rotate (x8) \\ 
 \hline
 ZJU231 & 56.72 & 0.99752 & 0.17 & GTX 1080 Ti & self ensemble \\  
 \hline
 NoahDn & 56.47 & 0.99749 & 3.54 & Tesla V100 & flip/rotate (x8)  \\     
 \hline
 Dahua\_isp & 56.20 & 0.99749 & ? & GTX 2080 & ?  \\  
 
 \hline\end{tabular}

\end{table*}

\subsection{Comparison to State-of-the-Art}
In this section, we compare our method's performance to the state-of-the-art on SIDD+ Test Set.
We have evaluated the PSNR and MS-SSIM metrics of our method on a copied evaluation server of NTIRE-2020 Denoising Challenge - rawRGB Track, thanks to the original challenge organizers.
We have evaluated two variants of our method with SDWT bottleneck: (a) 8 SDWT blocks and 64 filters ($SDWT_{8,64}$) (b) 12 SDWT blocks and 96 filters ($SDWT_{12,96}$).
As the goal of the challenge was not perceptual quality, but distortion metrics, for these evaluations, we have only used L1 loss in order to aim for the best PSNR for each variant.

\begin{figure*}
  \centering \includegraphics[scale=0.45]{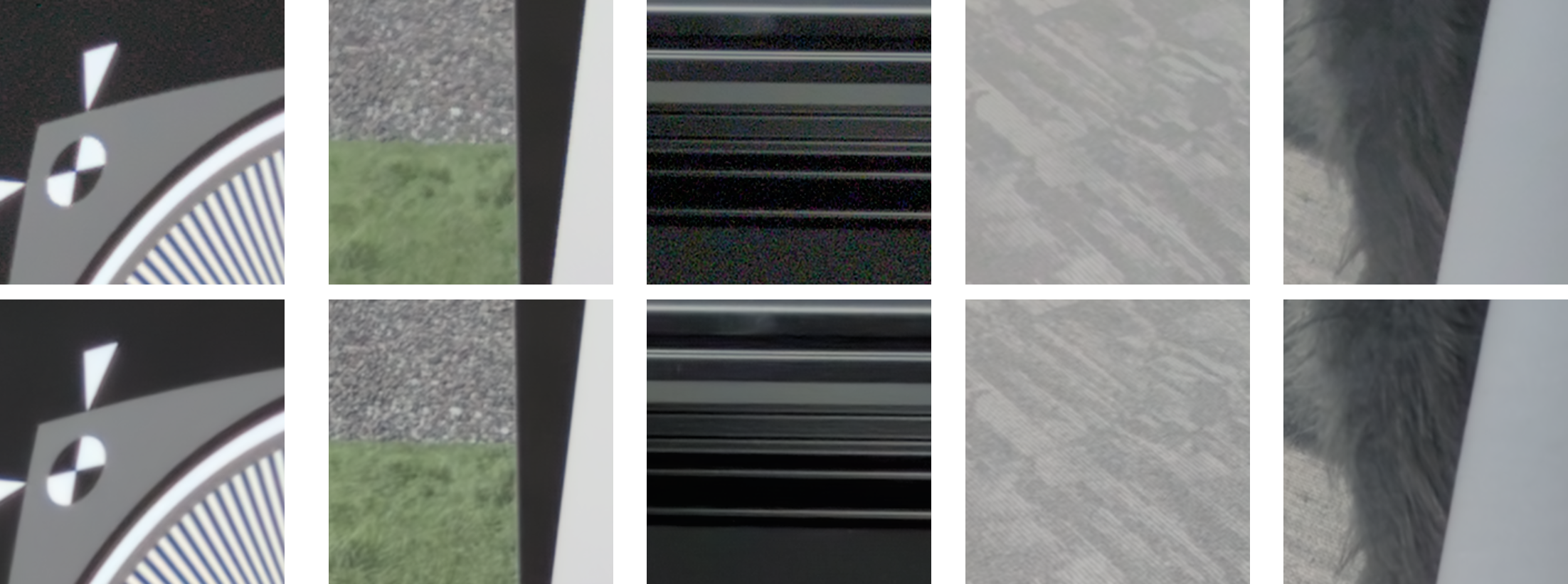}
  \caption{Visual Comparisons: NLM (top) vs $SDWT_{top-k}$ (bottom)}
  \label{nlmcomp}
\end{figure*}

The relevant information in Table \ref{test_psnr} are directly copied from \cite{NTIRE2020}. 
Our heavier model $SDWT_{12,96}$ model ranks among the top-performing methods, whereas our model $SDWT_{8,64}$ gives a good speed and PSNR compromise. 
We would like to note here that the run-time of our method is calculated by considering the ensembles, for example $SDWT_{12,96}$ operates at 6.63 seconds per MPixel on 8 ensembles, which means per ensemble it runs with 0.83 seconds per MPixel.

\subsection{Experiments on Our Own Camera Modules} \label{own_exps}
In order to fully investigate the method in real use case, we have also conducted experiments in our own camera modules.
In these experiments, the only difference from above experimental settings is that we have simulated the noisy data from SIDD ground truths according to our module's noise model.
Although we do not make use of the real noisy images in SIDD, the dataset is still irreplaceable due to the noise-free (ground truth) images it provides, which enables us to create realistically simulated noisy images.
Note that in all experiments in this section, we have used our own ISP to generate images for visualization.

As the most important criterion for camera manufacturers, we have focused on visual quality improvement in these experiments.
We first visually compare our method $SDWT_{top-k}$ to a baseline method NLM \cite{NLM}. NLM was algorithmically adopted to process Bayer data and signal-dependent noise model, and tuned for best performance in our cameras.
The comparison is made in test images captured with our camera at lab and in real world.
It can be observed from Fig. \ref{nlmcomp} that $SDWT_{top-k}$ clearly outperforms NLM in representative samples from many cases: flat areas, dense lines and various textures.

\begin{figure}
  \centering
  \subfigure[$SDWT$]{\label{siemens1}\includegraphics[width=0.21\textwidth]{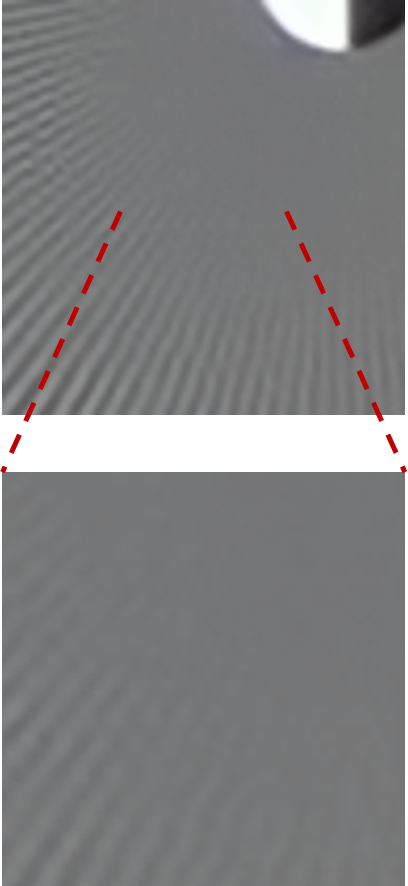}}\qquad
  \subfigure[$SDWT_{top-k}$]{\centering\label{siemens2}\includegraphics[width=0.21\textwidth]{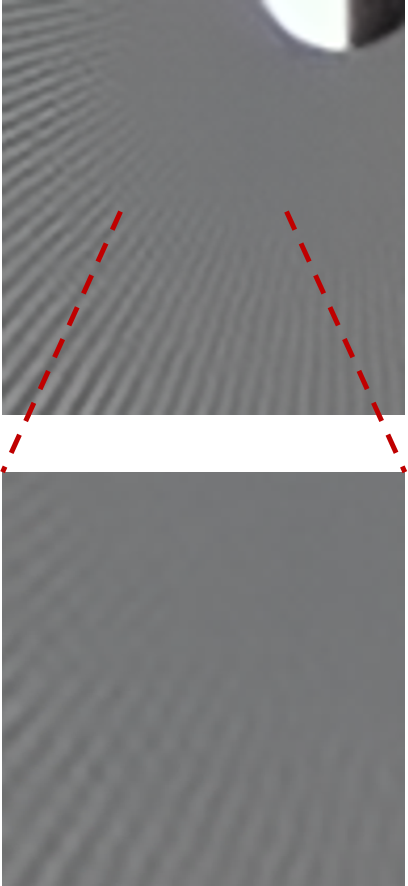}}
\caption{Visual comparison on Siemens Star patches from TE-42 Chart: Top: Patches from gray Siemens Star, Bottom: Enlarged middle-crops from the patches}
\label{siemens}
\end{figure}

We also compare $SDWT$ to $SDWT_{top-k}$ to highlight the differences visual quality that our proposed loss function brings.
%Although we have verified the validity of visual quality improvement of $SDWT_{top-k}$ over $SDWT$ in other images, here we only share results on a cropped Siemens-Star from TE-42 chart due to confidentiality.
As it can be observed from Fig. \ref{siemens}, $SDWT_{top-k}$ in Fig. \ref{siemens2} provides a good restoration of the Siemens star lines whereas $SDWT$ in Fig. \ref{siemens1} over-smoothens the image, especially when lines gets denser. 
This shows that $SDWT_{top-k}$ indeed preserves more details which is an important factor for perceptual quality.

\section{Conclusion}
We have proposed two contributions to deep neural networks based denoising.
We have shown that applying operations on each sub-band separately increases the accuracy of denoising in terms of both visual quality and distortion metrics.
We argue that this is due to keeping the sub-band order uncorrupted prior to IDWT as opposed to common approach which applies operations on concatenated sub-bands, hence potentially corrupting the sub-band order.
Our second contribution, the top-$k\%$ loss was shown to largely keep the PSNR performance while notably improving the visual quality.
We argue that this is due to enabling a denoised image that is, in each location, frequency-wise more balanced in terms of the error in DCT coefficients.

{\small
\bibliographystyle{ieee_fullname}
\bibliography{egbib}

\begin{thebibliography}{10}\itemsep=-1pt

\bibitem{NTIRE2020}
Abdelrahman Abdelhamed, Mahmoud Afifi, Radu Timofte, and Michael~S Brown.
\newblock Ntire 2020 challenge on real image denoising: Dataset, methods and
  results.
\newblock In {\em Proceedings of the IEEE/CVF Conference on Computer Vision and
  Pattern Recognition Workshops}, pages 496--497, 2020.

\bibitem{SIDD}
Abdelrahman Abdelhamed, Stephen Lin, and Michael~S Brown.
\newblock A high-quality denoising dataset for smartphone cameras.
\newblock In {\em Proceedings of the IEEE Conference on Computer Vision and
  Pattern Recognition}, pages 1692--1700, 2018.

\bibitem{NTIRE2019}
Abdelrahman Abdelhamed, Radu Timofte, and Michael~S Brown.
\newblock Ntire 2019 challenge on real image denoising: Methods and results.
\newblock In {\em Proceedings of the IEEE/CVF Conference on Computer Vision and
  Pattern Recognition Workshops}, pages 0--0, 2019.

\bibitem{SIDDBENCH}
{Abdelrahman Abdelhamed}.
\newblock Sidd benchmark.
\newblock \url{https://www.eecs.yorku.ca/~kamel/sidd/benchmark.php}.
\newblock [Online; accessed 13-February-2021].

\bibitem{RENOIR}
Josue Anaya and Adrian Barbu.
\newblock Renoir-a benchmark dataset for real noise reduction evaluation.
\newblock {\em Journal of Visual Communication and Image Representation}, pages
  144--154, 2018.

\bibitem{MSRD}
Long Bao, Zengli Yang, Shuangquan Wang, Dongwoon Bai, and Jungwon Lee.
\newblock Real image denoising based on multi-scale residual dense block and
  cascaded u-net with block-connection.
\newblock In {\em Proceedings of the IEEE/CVF Conference on Computer Vision and
  Pattern Recognition Workshops}, pages 448--449, 2020.

\bibitem{PERDIS}
Yochai Blau and Tomer Michaeli.
\newblock The perception-distortion tradeoff.
\newblock In {\em Proceedings of the IEEE Conference on Computer Vision and
  Pattern Recognition}, pages 6228--6237, 2018.

\bibitem{NLM}
Antoni Buades, Bartomeu Coll, and J-M Morel.
\newblock A non-local algorithm for image denoising.
\newblock In {\em 2005 IEEE Computer Society Conference on Computer Vision and
  Pattern Recognition (CVPR'05)}, volume~2, pages 60--65. IEEE, 2005.

\bibitem{BM3D}
Kostadin Dabov, Alessandro Foi, Vladimir Katkovnik, and Karen Egiazarian.
\newblock Image denoising by sparse 3-d transform-domain collaborative
  filtering.
\newblock {\em IEEE Transactions on image processing}, 16(8):2080--2095, 2007.

\bibitem{KSVD}
Michael Elad and Michal Aharon.
\newblock Image denoising via sparse and redundant representations over learned
  dictionaries.
\newblock {\em IEEE Transactions on Image processing}, 15(12):3736--3745, 2006.

\bibitem{WNNM}
Shuhang Gu, Lei Zhang, Wangmeng Zuo, and Xiangchu Feng.
\newblock Weighted nuclear norm minimization with application to image
  denoising.
\newblock In {\em Proceedings of the IEEE conference on computer vision and
  pattern recognition}, pages 2862--2869, 2014.

\bibitem{RESNET}
Kaiming He, Xiangyu Zhang, Shaoqing Ren, and Jian Sun.
\newblock Deep residual learning for image recognition.
\newblock In {\em Proceedings of the IEEE conference on computer vision and
  pattern recognition}, pages 770--778, 2016.

\bibitem{DENSENET}
Gao Huang, Zhuang Liu, Laurens Van Der~Maaten, and Kilian~Q Weinberger.
\newblock Densely connected convolutional networks.
\newblock In {\em Proceedings of the IEEE conference on computer vision and
  pattern recognition}, pages 4700--4708, 2017.

\bibitem{JPEG}
Graham Hudson, Alain L{\'e}ger, Birger Niss, Istv{\'a}n Sebesty{\'e}n, and
  J{\o}rgen Vaaben.
\newblock Jpeg-1 standard 25 years: past, present, and future reasons for a
  success.
\newblock {\em Journal of Electronic Imaging}, 27(4):040901, 2018.

\bibitem{GRDN}
Dong-Wook Kim, Jae Ryun~Chung, and Seung-Won Jung.
\newblock Grdn: Grouped residual dense network for real image denoising and
  gan-based real-world noise modeling.
\newblock In {\em Proceedings of the IEEE/CVF Conference on Computer Vision and
  Pattern Recognition Workshops}, pages 0--0, 2019.

\bibitem{BAUG}
Jiaming Liu, Chi-Hao Wu, Yuzhi Wang, Qin Xu, Yuqian Zhou, Haibin Huang, Chuan
  Wang, Shaofan Cai, Yifan Ding, Haoqiang Fan, et~al.
\newblock Learning raw image denoising with bayer pattern unification and bayer
  preserving augmentation.
\newblock In {\em Proceedings of the IEEE/CVF Conference on Computer Vision and
  Pattern Recognition Workshops}, pages 0--0, 2019.

\bibitem{RADAM}
Liyuan Liu, Haoming Jiang, Pengcheng He, Weizhu Chen, Xiaodong Liu, Jianfeng
  Gao, and Jiawei Han.
\newblock On the variance of the adaptive learning rate and beyond.
\newblock {\em arXiv preprint arXiv:1908.03265}, 2019.

\bibitem{MLWCNN}
Pengju Liu, Hongzhi Zhang, Kai Zhang, Liang Lin, and Wangmeng Zuo.
\newblock Multi-level wavelet-cnn for image restoration.
\newblock In {\em Proceedings of the IEEE conference on computer vision and
  pattern recognition workshops}, pages 773--782, 2018.

\bibitem{DSGW}
Wei Liu, Qiong Yan, and Yuzhi Zhao.
\newblock Densely self-guided wavelet network for image denoising.
\newblock In {\em Proceedings of the IEEE/CVF Conference on Computer Vision and
  Pattern Recognition Workshops}, pages 432--433, 2020.

\bibitem{SKNAS}
Marcin Mozejko, Tomasz Latkowski, Lukasz Treszczotko, Michal Szafraniuk, and
  Krzysztof Trojanowski.
\newblock Superkernel neural architecture search for image denoising.
\newblock In {\em Proceedings of the IEEE/CVF Conference on Computer Vision and
  Pattern Recognition Workshops}, pages 484--485, 2020.

\bibitem{TDWT}
Quan Pan, Lei Zhang, Guanzhong Dai, and Hongai Zhang.
\newblock Two denoising methods by wavelet transform.
\newblock {\em IEEE transactions on signal processing}, 47(12):3401--3406,
  1999.

\bibitem{DCHN}
Bumjun Park, Songhyun Yu, and Jechang Jeong.
\newblock Densely connected hierarchical network for image denoising.
\newblock In {\em Proceedings of the IEEE/CVF Conference on Computer Vision and
  Pattern Recognition Workshops}, pages 0--0, 2019.

\bibitem{TID}
Nikolay Ponomarenko, Lina Jin, Oleg Ieremeiev, Vladimir Lukin, Karen
  Egiazarian, Jaakko Astola, Benoit Vozel, Kacem Chehdi, Marco Carli, Federica
  Battisti, et~al.
\newblock Image database tid2013: Peculiarities, results and perspectives.
\newblock {\em Signal processing: Image communication}, 30:57--77, 2015.

\bibitem{UNET}
Olaf Ronneberger, Philipp Fischer, and Thomas Brox.
\newblock U-net: Convolutional networks for biomedical image segmentation.
\newblock In {\em International Conference on Medical image computing and
  computer-assisted intervention}, pages 234--241. Springer, 2015.

\bibitem{DIDU}
Songhyun Yu, Bumjun Park, and Jechang Jeong.
\newblock Deep iterative down-up cnn for image denoising.
\newblock In {\em Proceedings of the IEEE/CVF Conference on Computer Vision and
  Pattern Recognition Workshops}, pages 0--0, 2019.

\bibitem{RDB}
Yulun Zhang, Yapeng Tian, Yu Kong, Bineng Zhong, and Yun Fu.
\newblock Residual dense network for image restoration.
\newblock {\em IEEE Transactions on Pattern Analysis and Machine Intelligence},
  2020.

\bibitem{EPLL}
Daniel Zoran and Yair Weiss.
\newblock From learning models of natural image patches to whole image
  restoration.
\newblock In {\em 2011 International Conference on Computer Vision}, pages
  479--486. IEEE, 2011.

\end{thebibliography}
}

\end{document}